\documentclass[letter, 10 pt, conference]{ieeeconf}  

\usepackage{epsfig}
\usepackage{epstopdf}
\usepackage{amsmath}
\usepackage{amssymb}
\usepackage{bm}
\usepackage{graphicx} 
\usepackage{booktabs}

\usepackage[font=footnotesize,skip=4pt]{subcaption}
\usepackage[font=footnotesize,skip=2pt]{caption}

\usepackage{algorithm}
\usepackage{algorithmic}
\usepackage{graphicx}
\usepackage{subcaption}
\usepackage{caption}
\usepackage{balance}

\IEEEoverridecommandlockouts                              

\usepackage{graphics} 
\usepackage{epsfig} 
\usepackage{mathptmx} 
\usepackage{times} 
\usepackage{amsmath} 
\usepackage{amssymb}  
\usepackage[usenames, dvipsnames]{color}

\title{\large \textbf{Color-Coded Fiber-Optic Tactile Sensor for an Elastomeric Robot Skin}
}

\author{Zhanat Kappassov, \IEEEmembership{Member, IEEE}, Daulet Baimukashev, Zharaskhan Kuanyshuly, \IEEEmembership{Student Members, IEEE}, \\  Yerzhan Massalin, Arshat Urazbayev and Huseyin Atakan  Varol, \IEEEmembership{Senior Member, IEEE} 
	\thanks{This work was partially supported by  Nazarbayev University Faculty-development competitive research grants program  \lq\lq Variable Stiffness Tactile Sensor for Robot Manipulation and Object Exploration'' 110119FD45119 and the Ministry of Education and Science of the Republic of Kazakhstan grant \lq\lq Methods for Safe Human Robot Interaction with Variable Impedance Actuated Robots''.}
	\thanks{Z. Kappassov,  D. Baimukashev, Z. Kuanyshuly,  Y. Massalin, A. Urazbayev and H. A. Varol are with the Dept. of Robotics and Mechatronics, Nazarbayev University, 53 Kabanbay Batyr Ave, Z05H0P9 Astana, Kazakhstan. Email: \{zhkappassov, dbaimukashev, zharaskhan.kuanyshuly, yerzhan.massalin, aurazbayev, ahvarol\}@nu.edu.kz.}
	\thanks{Corresponding author: Z. Kappassov.}
}

\begin{document}
	
	\bstctlcite{IEEEexample:BSTcontrol} 
	
	\maketitle
	\thispagestyle{empty}
	\pagestyle{empty}
	\begin{abstract}
%
%
%
%
%
%
%
%
The sense of touch is essential for reliable mapping between the environment and a robot which interacts physically with objects. Presumably, an artificial tactile skin would facilitate safe interaction of the robots with the environment. In this work, we present our color-coded tactile sensor, incorporating plastic optical fibers (POF), transparent silicone rubber and an off-the-shelf color camera. Processing electronics are placed away from the sensing surface to make the sensor robust to harsh environments. Contact localization is possible thanks to the lower number of light sources compared to the number of camera POFs. Classical machine learning techniques and a hierarchical classification scheme  were used for contact localization. Specifically, we generated the mapping from stimulation to sensation of a robotic perception system using our sensor. We achieved a force sensing range up to 18 N with the force resolution of around 3.6~N and the spatial resolution of 8~mm. The color-coded tactile sensor is suitable for tactile exploration and might enable further innovations in robust tactile sensing.
	\end{abstract}
\section{Introduction}

Somatosensory system is responsible for human sensations through the skin. It can be considered as the hidden ingredient of human dexterity in object manipulation. With the robots replacing humans in more and more tasks, tactile sensing is becoming essential for robots. This sense provides a direct mapping of the object's shape to the acquired signals~\cite{Terekhov20151661}. Such mapping is necessary both for dexterous manipulation and safe human-robot interaction (HRI)~\cite{Cherubini}. Development of robust and effective tactile sensors is a major challenge that could lead to advances in terms of the safety, responsivity, and dexterity of robots~\cite{kappassov_tactile_2015,LUO201754}. 

While sensing capabilities of the tactile sensors are improving~\cite{cherrier_tactip_2018}, there is also a need for qualitative improvements -- the sensors should be abrasion and water resistant -- as robots move from industrial workcells into human-inhabited environments~\cite{gates2007robot}. In human environments, a robot should be able to manipulate an object with liquids in it~\cite{Water_or_not} or do it immersed inside water~\cite{ocean}. Since water is an electroconductive media, a robust sealing is necessary to keep the electronic processing units safe. Sealing requires continuous maintenance and complex manufacturing together with proper assembly of the robot fingertips~\cite{KAMPMANN2015115}. This can be mitigated by placing fragile signal processing electronics away from the sensing unit, that is attached onto the fingertips. Such implementation would also allow a humanoid robot to have a centralized processing unit, e.g. in its torso.   

\begin{figure}[!b]
	\centering
	\includegraphics[width=1\columnwidth]{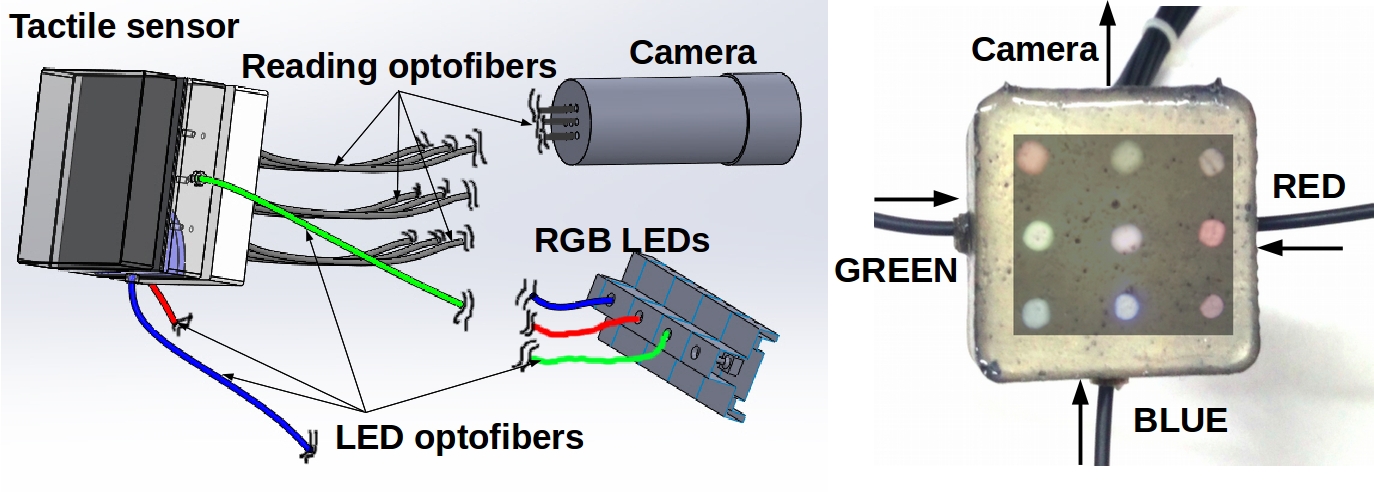}
	\vspace{-5 mm}
	\caption{Color-coded tactile sensor. (a) Assembly with LEDs, camera and plastic optical fibers. (b) Camera snapshot overlaid onto the sensor. The camera image is processed to infer the sensor deformation.}
	\label{fig:sensor_stack_image}
\end{figure}

We address this problem by developing an optical tactile sensing array, which uses physical interaction with an object to provide force sensing and contact localization. We deliver the light from three sources with different colors to a commodity vision camera via plastic optical fibers (POFs) embedded inside transparent silicone. We use this compressible silicone as a color-coded tactile sensor (see Fig.~\ref{fig:sensor_stack_image}). Our tactile sensor is energy efficient thanks to the use of less number of light sources than the camera POFs. Moreover, working with light beams rather than with a flow of electrons can be advantageous for some applications. For instance, data flow through electrical cables can be distorted by a magnetic field but magnetic fields do not interfere with optical signals~\cite{Kaspar_mrt}. 

In the remainder of this paper, we firstly review the related work in Sec.~\ref{sec:related}. In Section~\ref{sec:concept}, we explain the working principle and software development of our color-coded tactile sensor, and then we introduce its fabrication process in Sec.~\ref{sec:fabrication}. Afterward, we present our experimental results on force measurement and contact localization to benchmark the performance of our sensor (see Sec.~\ref{sec:characterization}). For these measurements, we use machine learning to measure the contact force from the previously acquired samples (see Sec.~\ref{sec:ML}). The last section summarizes our work.

\section{Related work}
\label{sec:related}
Several types of tactile sensors based on various transduction methods (e.g. capacitive, resistive, magnetic, and optical) were reviewed in~\cite{kappassov_tactile_2015}. Optical sensing has several advantages including high spatial resolution, sensitivity, repeatability and also immunity from electromagnetic interference.  Moreover, some of these optical sensors can work under high ambient pressures (up to 600 bar)~\cite{Kampmann_2014_Integration_optical_combined_F_Vibr}. Next, we first summarize the materials used in these optical tactile sensors and then overview the existing sensors.

\subsection{Materials}
\label{sec:materials}
There are different types of materials used in manufacturing of optical sensors: various polymers including silicone, polyurethane, and thermoplastic elastomers; POFs and hydrogels. 

Liquid silicone rubber compounds (e.g. \textregistered Smooth-on Sorta Clear 18 and \textregistered Techsil RTV27905) are widely used in injection molding to create robust parts. The part quality mainly depends on how well the silicone compounds are mixed during molding. On the other hand, thermoplastic rubbers, as in~\cite{s17122762}, provide a better ability to return to their original shape after stretching them to moderate elongations. These can be processed by heating the granules of the thermoplastic elastomer, shaping them under pressure, and then cooling them to solidify. In contrast to silicone rubber and elastomers, polyurethanes can be synthesized by chemical reactions. Polyurethane parts are resistant to wear and tear. 

Molding the above-mentioned materials requires manufacturing steps that can be avoided by 3D-printing. Stereolithography resins, e.g. \textregistered Formlabs or \textregistered Objet, can be used for rapid prototyping. However, these parts still lack the durability and strength of the classical counterparts.

With the aim of making wearable and biocompatible parts, technological advances in bioengineering led to the emergence of hydrogels~\cite{s100504381}. A hydrogel, a rubbery and transparent material composed mostly of water, can also be a good choice for safe physical HRI.

Robot sensing technology also leverages inventions in telecommunications besides the ones in bioengineering. Glass and plastic fibers can be used as force sensors~\cite{tossi_italy}. Stretchable fibers~\cite{stretch_light} that respond to deformations with a color variation can measure strain forces. Tactile sensing using such stretchable fibers is accomplished by periodically changing the refractive index along the length of a fiber.
\subsection{Optical Sensors}
\label{sec:opt_sensors}
Using the materials described in Sec~\ref{sec:materials}, a variety of optical tactile sensors were presented in the literature. 
The general principle is based on the optical reflection between mediums with different refractive indices. A conventional optical tactile sensor consists of an array of infrared light-emitting diodes (LEDs) and photodetectors. The intensity of the light is usually proportional to the magnitude of the pressure~\cite{tossi_italy}. 

GelSight tactile sensor~\cite{s17122762} uses a thermoplastic elastomer coated with a reflective membrane highlighted by an LED ring to capture surface textures with a camera. In~\cite{Shan}, this sensor was benchmarked in a texture recognition problem. Similarly, researchers of the Bristol Robotics laboratory developed a family of optical tactile sensors that are almost ready for small-scale mass production~\cite{cherrier_tactip_2018}. Their TacTip sensor uses a commodity image tracker originally used in optical computer mice. It combines an image acquisition system and a digital signal processor, capable of processing the images at 2000~Hz~\cite{Lepora}. Thanks to the high image processing rate, they can detect the slippage of a grasped object~\cite{Lepora_slip}. In~\cite{ohka_takata_kobayashi_suzuki_morisawa_yussof_2009}, a touch sensor, consisting of 41  silicone rubber markers, a light source, and a camera, estimates the tangential and normal forces by tracking these markers. Markers with different colors are used in GelForce sensor~\cite{5306070}.

Researchers embedded an optical tactile sensor into the multi-modal tactile sensing system of an underwater robot gripper~\cite{Kampmann_2014_Integration_optical_combined_F_Vibr}. As in~\cite{Kaspar_mrt} and \textregistered {Optoforce sensor}, the sensing principle is based on the light reflection delivered via POFs. The POFs can be used as force sensing elements due to the stray light, which is considered as a drawback in telecommunications~\cite{tossi_italy}. The deformation of a POF increases the losses of the light propagated inside, as the attenuation coefficient increases. In additon, the elasto-optic metamaterial presented in~\cite{stretch_light} can change its refractive index due to pure bending. Such POFs are fabricated by the chemical vapor deposition technique. Their design generally relies upon the phenomenon of optical interference~\cite{Nagano:78}.

Laboratory prototypes of image-based tactile sensors were reported in~\cite{Saga_reflection} and~\cite{5306070}. In these sensing panels, LEDs and photo-diodes/camera were placed against a reflecting planar surface. When the surface deforms, it causes changes in reflected beams. These sensors use optical light to detect deformation of the contact surface, which can be used to estimate the force.

In contrast to the above-mentioned approaches, we aim to keep the processing and conditioning electronics away from the sensing matter to increase the durability. In order to estimate both the contact force and its location, we feed the sensor with three different colors. The details of our solution are covered in the next sections.

\section{Color-Coded Sensor Design}
\label{sec:concept}

We tackled the issues in the manufacturing and signal conditioning, that the current sensors share to some extent, using off-the-shelf components. To enable force measurements, a commodity camera records the light intensity of POFs embedded in a layer of silicone (Smooth-On Sorta Clear 18). Transparent silicone is widely used in molding applications. It does not require complex manufacturing, while satisfying the transparency and mechanical requirements, such as water and abrasion resistance. In the following, we first describe the sensing principle using such silicone with a commodity camera and then describe our sensor fabrication process.

\subsection{Sensing Principle}
\begin{figure}[]
	\centering
	\vspace{3 mm}
	\includegraphics[width=0.45\textwidth]{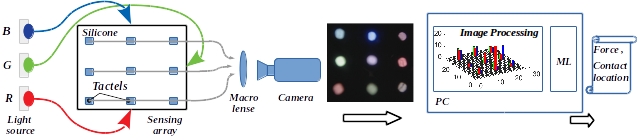}
	\caption{The block diagram illustrating the sensing principle of the color-coded optical tactile sensor.} 
	\label{fig:proper_principle_flow_chart}
\end{figure}

Figure~\ref{fig:proper_principle_flow_chart} illustrates the optical signal processing pipeline. Light sources illuminate a sensing silicone. The illumination is guided via POFs. A camera, incorporating a macro lens, captures the light illumination trapped in the silicone. A computer processes the recorded signals, and a machine learning algorithm classifies the processed data. As a result we can obtain the location and normal force when the sensor is in contact with the environment. We choose a transparent silicone as the light propagation medium. This silicone is supplied by three light emitting diodes (LEDs) with different colors. Plastic optical fibers deliver this light into the silicone material and also deliver the emerging color within this matter to a commodity red-green-blue (RGB) vision sensor. We utilize twelve POFs: three of these POFs emit light into the silicone. The emitted light is scattered inside silicone and is received by the rest nine POFs. 

There are three light sources:  red, green and blue. When red, green and blue light frequency bands are mixed together in equal proportions, white color appears. If one of these bands is excluded, another color is observed, e.g. cyan when the red band is excluded.  In contrast, when the red band is more intense than green and blue, LightCoral ($\#F08080$) appears. The color changes can be detected by acquiring the color data using a RGB camera. When mixed in some soft and transparent matter, these color changes might be used to measure material deformations. 

\begin{figure}[]
	\centering
		\includegraphics[width=0.45\textwidth]{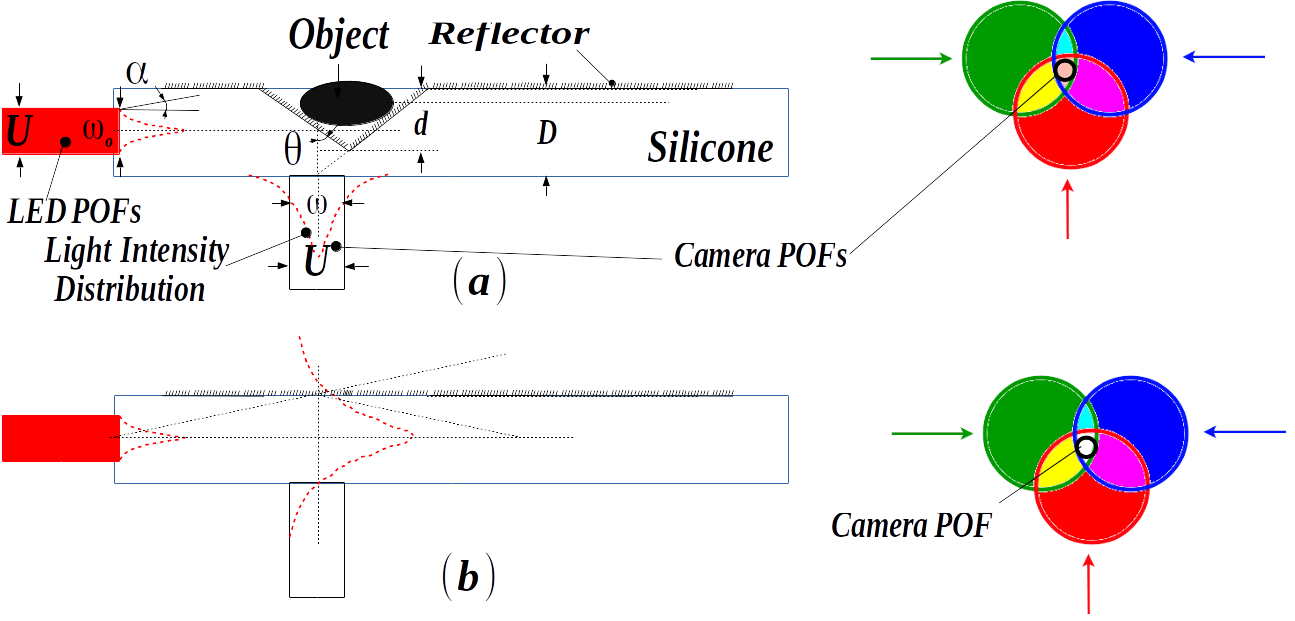}
	\caption{Sensing principle: (a)  change of the light intensity and color of a single POF due to deformation; (c) no change of the light inensity} 
	\label{fig:proper_principle}
\end{figure}
As soon as the silicone gets compressed after contact with an object, the light scattering pattern changes.  If its color changes, one can use the color chart to determine whether and to what depth the silicone is deformed. Thus, the color-coded silicone substrate acts as a pressure-sensing media and changes color to signal pressure level. A geometrical explanation of this principle is shown in Fig.~\ref{fig:proper_principle}. The figure exemplifies a one-dimensional case with one light source, e.g. red, POF and one camera POF. A given POF can get more light than others and vice versa as the directions of the reflected light beams change during the deformation of the sensing surface under an external force. Since the silicone substrate can be approximated by a spring with a constant Young’s modulus~$E$, then this external force as function of deformation $d$ is given by $F \propto \frac{E A}{D} \ d,$ where $A$ is the contact area over the sensor and $D$ is the thickness of the silicone rubber~\cite{Correll_boulder_silicone_force}. Depending on the location of the deformation and its depth, the emitted light intensity, $I_0$, decreases with the beam path $I(r)\propto I_0 e^{-2\frac{r^2}{\omega^2}}$\cite{Kaspar_mrt}, where $I(r)$ is the measured reflected intensity, $r$ is the beam path to a camera POF, and $\omega_0$ and $\omega$ are the spot size of the Gaussian beam at the source $r=0$ and at the camera POF, respectively. In some cases, e.g. Fig.~\ref{fig:proper_principle}(a), $r$ is shorter than when the sensor is not in contact, e.g. Fig.~\ref{fig:proper_principle}(b). The light is trapped inside the silicone thanks to the reflection that is realized using a thin film. The angle of the reflection, $\theta$ modifies this path too. Moreover, the compression of the silicone eventually alters the scattering and absorption properties, which dramatically increase for large deformations. For example, the LED light intensity decreases due to the increased density of irregularities. This effect depends on many properties such as the chemical composition and impurities, the details of which are beyond the scope of this work. However, we should note that $I_0$ is heavily affected by absorption when the silicone is compressed.
\begin{figure}[!b]
	\centering
	\includegraphics[width=1\columnwidth]{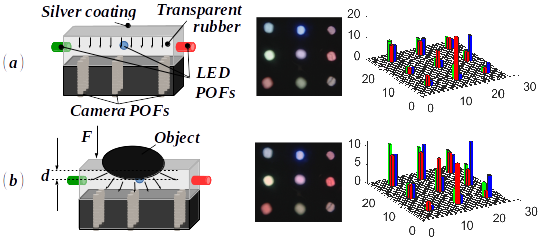}
	\caption{Change of color in our color-coded tactile sensor. Depending on an external force, the silicone deforms and the colors acquired by the camera via POFs changes. (a) External force is zero. (b) Light intensities change as the sensing surface deforms under applied external force.}
	\label{fig:sensor_press}
\end{figure}
\subsection{Raw Data Descriptor}
\label{sec:data_proc}
In this subsection, we explain the image processing for our sensor. Since the acquired signals are light intensities; the change of red, green, and blue channel intensities of each reflected fiber can be used as a descriptor to infer both the force applied to the sensor array and the location of the contact. The descriptor algorithm is implemented using Robot Operating System (ROS) and OpenCV library. First, we acquire the images at 15 Hz with $640\times480$ resolution. Then for each image, we define the regions of interest (ROI)~$R$ around the locations of every POF. As a result, for the input images, we define the ROIs regions $R_i,~i=1,2...9 \in Z^{w\times h}$, where $w$ and $h$ denote the ROI width and height, respectively. In each region, we exclude the black pixels and normalize the red, green and blue pixel values. We get nine mean values for each channel resulting in 27 features $\mu_{j},~j=1,2...27$ representing the surface deformations. Figure~\ref{fig:sensor_press} illustrates this descriptor which is used to obtain the feature vector. Using the resultant feature vector $s = (\mu_{1},\mu_{2}, \mu_{3}, ..., \mu_{26}, \mu_{27}) \in R^{27}$, we infer the sensing surface deformation as described in Sec.~\ref{sec:ML}. Any surface deformation depends on the sensor material and fabrication which is described in the following section. 
%
%
\section{Fabrication}
\label{sec:fabrication}
\begin{figure}[b]
	\centering
	\includegraphics[width=0.6\linewidth]{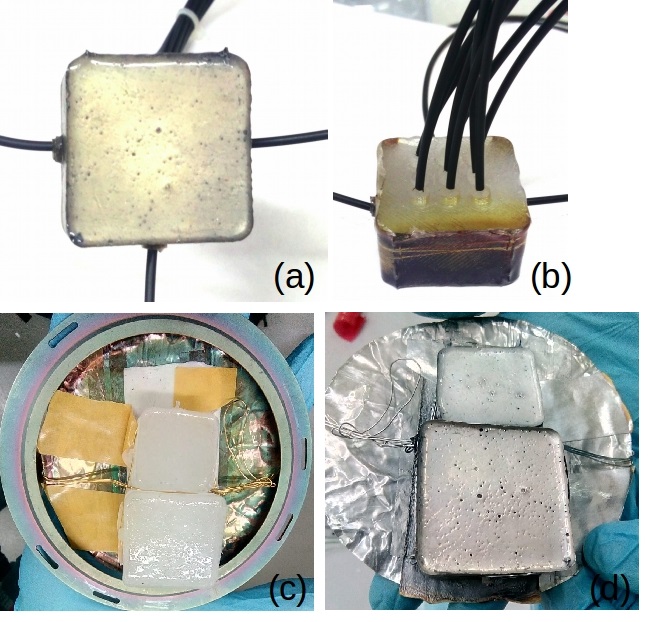}
	\caption{Color-coded tactile sensor: (a) Front view of the sensor with silver thin film. (b) Entrance of the optofibers and their sockets into the sensor. (c) Silicone without the silver thin film. (d) Silicone with the silver thin film.}
	\label{fig:sensor_silicone_fibers}
\end{figure}

Eventually, we aim to attach the color-coded tactile sensor to robot grippers for object manipulation. Therefore, the size of our sensor is $40 \times 40 \times 5~mm$. There are three fiber connections from the sides and nine fiber connections from the bottom. Figure~\ref{fig:sensor_silicone_fibers}(a,b) illustrate the fabricated sensor with the attached POFs. The distance between two neighboring POFs at the bottom side is 5 mm in both directions. The fibers that are on the sides are placed at the center of the corresponding edges. The depth of the inserted fibers is 5 mm. There are also extruded adapter sockets for the POFs. These sockets with the height of 8 mm increase the robustness of the connection between the POFs and silicone. The shape of the sensor is obtained by molding silicone into a plastic mold. The mold is 3D-printed using PLA filament. The structure of the sensor includes three layers: translucent silicone, reflective thin film, and silicone with injected reflecting particles for durability and better light reflection.       
\subsection{Silicone Sensor Layers}
\subsubsection{Silicone Compound}
The silicone is used both as the light transmission media and as a protective layer. There are different types of silicone compounds with  different shore durometers~\cite{Silicone_mechanical_property_Indentor}. The durometer scale determines the hardness of elastic materials. For our tactile sensor, we used Sorta Clear silicone compound with Shore level 18 from Smooth-On. We chose this material over the commonly used Polydimethylsiloxane (PDMS), e.g. as used in~\cite{Correll_boulder_silicone_force}, due to its better elasticity and durability. This allows large sensor surface deformations without sustaining damage~\cite{balloon}. 

The silicone is first poured into a mold. Then, the mold is degassed in a vacuum chamber. The proportion of the silicone components (part A and part B) is 10 to 1 and the solidification time is 24 hours. After the molded silicone cures, it is covered with a reflective thin film.
\subsubsection{Thin Film Magnetron Physical Vapor Deposition}
The effect of a thin film reflective layer is exemplified in Fig.~\ref{fig:sensor_silicone_fibers}(a). Figures~\ref{fig:sensor_silicone_fibers}(c) and ~\ref{fig:sensor_silicone_fibers}(d) illustrate a silicone sample before and after adding the thin film. We tried several materials including Aluminum (Al), Copper (Cu) and Silver (Ag) for making this thin film. For Al target, the obtained thin film was black (probably due to the fast oxidization). The reflectance characteristics of Cu was not desirable. Ag worked best for our purposes without the oxidization problem.

To create this thin film, we used magnetron physical vapor deposition machine (Kurt Lesker Lab-18). We installed the silicone sample in front of the Ag electrode and pumped the chamber for 90 minutes up to a pressure of $10^{-5}$~mT. The deposition of the Ag thin films on top of the silicone substrate is done by the magnetron sputtering at room temperature for eight minutes at 100 W. When the power is applied to the Ag electrode, the electrons accelerate in an electromagnetic field and the applied voltage ionizes the Ar gas. This leads to the formation of the plasma ring on the surface of the Ag electrode, which sputters the Ag atoms from the electrode and results in the formation of the Ag thin film on the surface of the silicone sensor. 

\subsubsection{Protective Layer}
The coated thin film wears and tears during physical contact with the environment. Therefore, we added one more silicone layer: Smooth On Sorta clear 18 mixed with a nickel powder. The nickel powder\footnote{Nickel Silver Powder Smooth-On(c) https://www.smooth-on.com/products/nickel-silver-metal-powder/} is added to increase the reflectance of the thin film as the surface of the silicone is not perfectly flat (see Fig.~\ref{fig:sensor_silicone_fibers}(d)).  

\subsection{System Integration}
\begin{figure}[!b]
	\centering
	\includegraphics[width=1\linewidth]{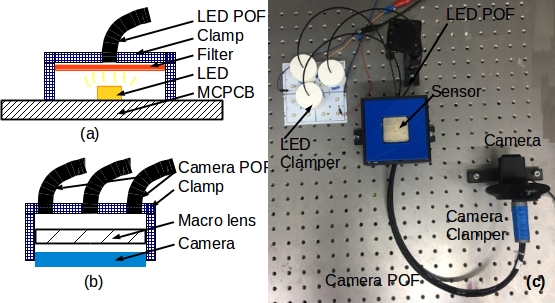}
	\caption{Clamps: (a)~Clamping structure for an LED, color filter and POF, (b)~clamping structure for the camera with the macro lens and POFs. (c)~System integration.}
	\label{fig:system_integration}
\end{figure}

The deformation of the sensor illuminated from the side via three plastic POFs is captured from the bottom by a commodity camera via nine more POFs as shown in Fig.~\ref{fig:system_integration}.   

\subsubsection{Light Sources}
We obtain red, green, blue light sources by use of corresponding color filters and high power surface-mount LEDs (Nichia NVSW219CT). These LEDs are soldered on a metal-core printed circuit board (MCPCB), and are driven with 85~mA current to obtain the luminous flux of 200~lm. A color filter is placed onto each LED (fixed by a custom 3D-printed adapter, which is also used as a clamp for the POFs as shown in Fig.~\ref{fig:system_integration}(a)). Figure~\ref{fig:system_integration}(c) illustrates how these clamps are placed on the MCPCB. There are 16 LEDs on it, but we only use three.
\subsubsection{Optofibers}
In order to prevent stray light, we used TLC 1 mm Simplex Plastic Fiber (Part Number: P96TB01TRBL22), which was covered with an insulation layer. Diameter of the core is 1~mm, and the external diameter with the insulation layer is 2.2~mm. Glue was used to fix the fibers to the LED clamps (see Fig.~\ref{fig:system_integration}(a)). In order to avoid a drop of the light intensity, the ends of the fibers were polished using a rotary tool (Dremel 4300) with fine sandpaper.  
\subsubsection{Camera}
The light beams from POFs are acquired to a desktop computer (Intel Xeon E5620, 4 GB DDR3 memory, Ubuntu 16.04 Linux operating system) using an off-the-shelf camera (Logitech HD Pro Webcam C920) at 15~Hz. Data is sent via Robot Operating System (ROS) environment at 15~Hz using \textit{libuvc} library. We replaced the lens of the camera with a macro-lens (2 cm focal length) due to the short distance between the sensing POFs and the camera. POFs are fixed so that the ends of them are facing the camera lens vertically (see Fig.~\ref{fig:system_integration}(b)). 
 
\section{Sensor Characterization}
\label{sec:characterization}

The fabricated sensor measures the contact force applied to its sensing surface. This contact force can be inferred from the changes in the camera image. Therefore, we need the relation between the applied force and ROI mean values for each color channel. We find this relation by compressing our tactile sensor using a robot arm equipped with a ground-truth force sensor. Exact relation between the applied force and changes in the camera images depends on a variety of fabrication properties, the details of which are beyond the scope of this paper. More details can be found in~\cite{pdms_materials} for light propagation and in~\cite{Silicone_mechanical_property_Indentor} for mechanical properties of silicone materials. 

\subsection{Platform and Experimental Scenario}
\label{sec:procedure}
\begin{figure}[!b]
	\centering
	\includegraphics[width= 1\columnwidth]{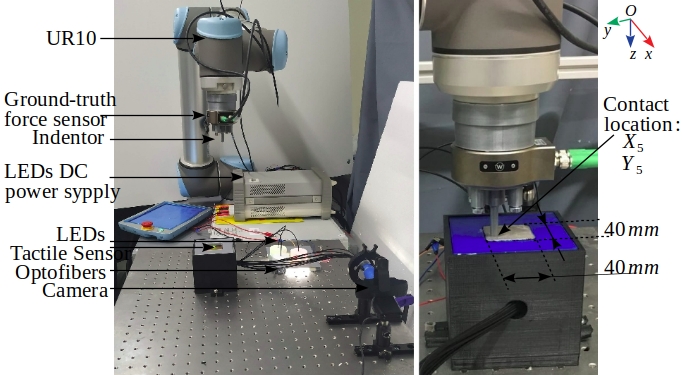}
	\caption{Calibration setup consisting of the UR10 robot arm (Universal Robots) and Weiss~KMS~40 force sensor with the attached indentor (left). The indentor pressing down on the silicone sensor (right).}
	\label{fig:calib_setup}
\end{figure}

We characterized the fabricated sensor using an industrial robot. The robot (UR10, Universal Robots) was controlled using the Position/Velocity command interface running at 125 Hz. Ground truth force measurements were obtained by a force/torque sensor (Weiss~KMS~40) attached to the robot's end-effector. The arm controller and the force sensor interfaces were both implemented using ROS. An indentor~\cite{Silicone_mechanical_property_Indentor} was attached on the ground-truth force sensor to compress the tactile sensor at exact locations. The indentor has a dome-shape with 3~mm tip radius. Figure~\ref{fig:calib_setup} shows this setup.

The robot end-effector moved vertically to press the color-coded tactile sensor, which was placed inside a 3D-printed plastic box and fixed on an optical table. First, we found the end-effector pose when the indentor was at the closest proximity to the sensor without registering any force. Then, the end-effector progressively squeezed the tactile sensor up to 3~mm with 0.6~mm steps (i.e. five depth levels). Afterward, the robot end-effector moved back until to the pre-touch position, positioned itself to a new location to squeeze the sensor again. There were 25 total locations, and therefore, $25\times5$ contact states. We recorded ten images at each contact state.

\subsection{Surface Deformation and Force}
\begin{figure}[!b]
	\centering
	\includegraphics[width=0.3\textwidth]{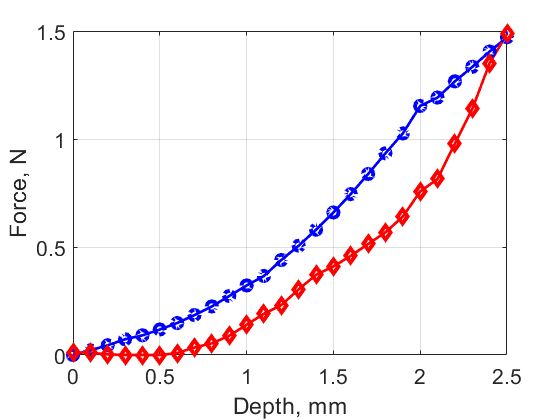}	
	\caption{Hysteresis: compression (blue) and release (red).  {Hysteresis~area}~$=~15$~Nm.}
	\label{fig:hysteresis}
\end{figure}
The thickness of the silicone dictates the maximum allowable displacement range and, thereby, the maximum measurement force and its resolution~\cite{Correll_boulder_silicone_force}. In our sensor design, we achieved the maximum detectable displacement range of 3~mm. Figure~\ref{fig:hysteresis} shows the response of the fabricated molded silicone up to this displacement and its hysteresis. Such displacement range would allow a robotic system to maintain robust contacts~\cite{kappassov_magnetic}. However, there is a trade-off between the maximum displacement range on one side, and the sensitivity and hysteresis on the other side.  

Force values and depth of the deformations exemplified in Fig.~\ref{fig:hysteresis} were obtained by squeezing the silicone at one of the 25 different locations. The locations of these contacts and the POFs are illustrated in Fig.~\ref{fig:calib_setup} right-hand side. The empirically derived Young's modulus of our sensor is $5.9$~MPa.  
\subsection{Estimation of Surface Deformation From the Image}
\label{sec:ML}
\begin{figure*}[]
	\centering
	\includegraphics[width=0.8\textwidth]{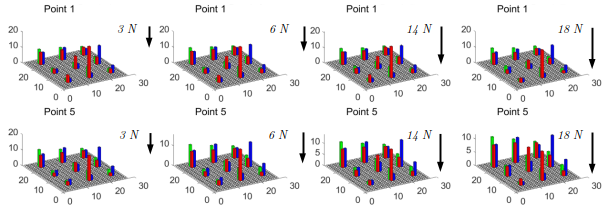}	
	\caption{3D surface visualization by feature vectors at different contact locations and depths: 1.2 mm (first column), 1.8 mm (second column), 2.4 mm (third column), and 3 mm (fourth column). Each feature vector corresponds to the applied contact force at a given contact location.}
	\label{fig:feature_vector_vs_depth}
\end{figure*}

As described in Sec.~\ref{sec:data_proc}, we obtain the feature vector $s$ by computing the mean values of ROIs. In order to validate this, we visualize this feature vector as bars with their lengths proportional to the elements of the vector. When plotted versus the surface deformation depths, the lengths of the bars vary for different contact states. Figure~\ref{fig:feature_vector_vs_depth} shows $s$ for two different locations Point 1 ($X_{1}, Y_{1}$) and Point 5 ($X_{5}, Y_{5}$) and four different depth levels ($Z_1, Z_2, Z_3, Z_{5}$).

The optical response of the sensor is not the same for every contact location. For instance, when the indentor squeezes our sensor at the location $(X_1, Y_1)$, the bars have lower heights than at the location $(X_5, Y_5)$. Thus, a  machine learning technique would be suitable to build the map from stimulation to sensation of a robotic perception system.

\subsection{Image-to-Force Map}
Various classification methods can be applied to estimate the contact information including the force~\cite{Lepora,s17122762}. We applied traditional machine learning techniques, such as Linear Discriminant Analysis (LDA), Quadratic Discriminant Analysis (QDA), Support Vector Machines (SVM) and $k-$nearest neighbors ($k-$NN).  In order to obtain the results, we collected  270 samples per class from 27 experiments (ten samples are recorded at every class) and allocated 20 of these experiments for training and seven for testing datasets. Number of classes is 125 due to 25 contact point locations and 5 depth levels of the indentor resulting in a 33750 sample dataset. The classification algorithms were implemented in Matlab with the default parameters. Table~\ref{table:results1} reports the mean accuracy obtained from the training dataset through 5-fold cross-validation and the testing dataset accuracy for generalization using the four chosen classification methods.
\begin{table}
	\begin{center}
		\caption{Image-to-force classification results}
		\begin{tabular}{ |c|c|c| }
			\hline
			Method & Cross-Validation Accuracy \% &  Generalization Accuracy \% \\
			\hline
			LDA & 76.1  & 73.6\\ 
			QDA & 89.8  & 80.0\\ 
			SVM & 90 & 80.96 \\ 
			$k-$nn & 91.8 & 81.5 \\ 
			\hline	
		\end{tabular}
		\label{table:results1}
	\end{center}
\end{table}

The classification methods have the generalization accuracy ranging from 73.6$\%$ to 81.5~$\%$. 

\subsection{Hierarchical Classification} 
Classification of all 125 classes representing the contact locations and depth levels in a single step results in the highest generalization accuracy of 81.55$\%$ using $k-$NN algorithm. In this case, the classification algorithm tries to localize and find the depth of contact simultaneously. We noted that the classification accuracy for contact localization only (25 classes) is much higher (99.2$\%$ for the testing dataset). 
\begin{figure}[!h]
	\centering
	\includegraphics[width=1.0\columnwidth]{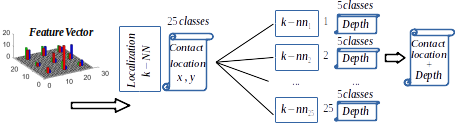}  
	\caption{Hierarchical localization of contact points.}
	\label{fig:ourcnn1}
\end{figure}

Thus, in order to infer the contact location and its depth from the images with more than 81.5$\%$ accuracy, we decided to sequentially find the contact location and then its depth. Figure~\ref{fig:ourcnn1} illustrates the resulting hierarchical classification pipeline. First, 20 random trials were selected for training. In the collected dataset, every contact location contains five depth levels. Each of these five depth levels were labeled as the same class to train the first level $k-$nn classifier to localize the contact. Then, for every contact location, a separate $k-$nn model was trained to retrieve the penetration depth of a given contact location. At this step, the five depth levels were labeled as a unique class. And we achieved 93.5$\%$ of accuracy for depth classification applied to the rest seven trials set aside for testing. As a result, the accuracy of finding the location and depth is increased to 92.7$\%$ with the hierarchical classification.

\section{Conclusion}
In this work, we presented the design and implementation of our POF-based tactile sensor. Our approach combines known optical design concepts with soft materials and utilizes three different colors to increase the spatial resolution. The sensor acts as a force-sensing media -- as its sensing surface deforms, the color acquired by the optical fibers changes. Such sensor provides several benefits that normally involve more complex designs (e.g. robustness and durability). We evaluate the design idea experimentally and benchmark the efficacy of our sensor using a machine learning-based approach.
Experimental results confirm that the color-coded tactile sensor is able to infer contact forces and their locations. Thus, our sensor has the potential to improve the dexterity of various robots in physical interaction tasks. In the future, we will improve our sensor design (by optimizing the design parameters) and also utilize it for tactile motion control (e.g. squeezing an object for determining its deformability).

\balance

\bibliographystyle{IEEEtran}
\bibliography{IEEEabrv,rgb_sensor____}

\begin{thebibliography}{10}
\providecommand{\url}[1]{#1}
\csname url@rmstyle\endcsname
\providecommand{\newblock}{\relax}
\providecommand{\bibinfo}[2]{#2}
\providecommand\BIBentrySTDinterwordspacing{\spaceskip=0pt\relax}
\providecommand\BIBentryALTinterwordstretchfactor{4}
\providecommand\BIBentryALTinterwordspacing{\spaceskip=\fontdimen2\font plus
\BIBentryALTinterwordstretchfactor\fontdimen3\font minus
  \fontdimen4\font\relax}
\providecommand\BIBforeignlanguage[2]{{%
\expandafter\ifx\csname l@#1\endcsname\relax
\typeout{** WARNING: IEEEtran.bst: No hyphenation pattern has been}%
\typeout{** loaded for the language `#1'. Using the pattern for}%
\typeout{** the default language instead.}%
\else
\language=\csname l@#1\endcsname
\fi
#2}}

\bibitem{Terekhov20151661}
A.~V. Terekhov and V.~Hayward, ``The brain uses extrasomatic information to
  estimate limb displacement,'' \emph{Proceedings of the Royal Society of
  London B: Biological Sciences}, vol. 282, no. 1814, 2015.

\bibitem{Cherubini}
B.~Navarro, \emph{et~al.}, ``In pursuit of safety: An open-source library for
  physical human-robot interaction,'' \emph{IEEE Robotics Automation Magazine},
  vol.~25, no.~2, pp. 39--50, June 2018.

\bibitem{kappassov_tactile_2015}
Z.~Kappassov, J.-A. Corrales, and V.~Perdereau,
  ``\BIBforeignlanguage{en}{Tactile sensing in dexterous robot hands:
  {Review}},'' \emph{\BIBforeignlanguage{en}{Robotics and Autonomous Systems}},
  vol.~74, pp. 195 -- 220, 2015.

\bibitem{LUO201754}
S.~Luo, \emph{et~al.}, ``Robotic tactile perception of object properties: A
  review,'' \emph{Mechatronics}, vol.~48, pp. 54 -- 67, 2017.

\bibitem{cherrier_tactip_2018}
B.~Ward-Cherrier, \emph{et~al.}, ``The {TacTip} {Family}: Soft optical tactile
  sensors with {3D}-printed biomimetic morphologies,'' \emph{Soft Robotics},
  vol.~5, no.~2, pp. 216--227, 2018.

\bibitem{gates2007robot}
B.~Gates, ``A robot in every home,'' \emph{Scientific American}, vol. 296,
  no.~1, pp. 58--65, 2007.

\bibitem{Water_or_not}
C.~Schenck and D.~Fox, ``Perceiving and reasoning about liquids using fully
  convolutional networks,'' \emph{The International Journal of Robotics
  Research}, vol.~37, no. 4-5, pp. 452--471, 2018.

\bibitem{ocean}
H.~Stuart, \emph{et~al.}, ``The ocean one hands: An adaptive design for robust
  marine manipulation,'' \emph{The International Journal of Robotics Research},
  vol.~36, no.~2, pp. 150--166, 2017.

\bibitem{KAMPMANN2015115}
P.~Kampmann and F.~Kirchner, ``Towards a fine-manipulation system with tactile
  feedback for deep-sea environments,'' \emph{Robotics and Autonomous Systems},
  vol.~67, pp. 115 -- 121, May 2015.

\bibitem{Kaspar_mrt}
H.~Xie, \emph{et~al.}, ``Magnetic resonance-compatible tactile force sensor
  using fiber optics and vision sensor,'' \emph{IEEE Sensors Journal}, vol.~14,
  no.~3, pp. 829--838, Mar 2014.

\bibitem{Kampmann_2014_Integration_optical_combined_F_Vibr}
P.~Kampmann and F.~Kirchner, ``Integration of fiber-optic sensor arrays into a
  multi-modal tactile sensor processing system for robotic end-effectors,''
  \emph{Sensors}, vol.~14, no.~4, pp. 6854--6876, 2014.

\bibitem{s17122762}
W.~Yuan, S.~Dong, and E.~H. Adelson, ``Gelsight: High-resolution robot tactile
  sensors for estimating geometry and force,'' \emph{Sensors}, vol.~17, no.~12,
  2017.

\bibitem{s100504381}
K.~Gawel, \emph{et~al.}, ``Responsive hydrogels for label-free signal
  transduction within biosensors,'' \emph{Sensors}, vol.~10, pp. 4381--4409,
  2010.

\bibitem{tossi_italy}
D.~Sartiano and S.~Sales, ``Low cost plastic optical fiber pressure sensor
  embedded in mattress for vital signal monitoring,'' \emph{Sensors}, vol.~17,
  no.~12, 2017.

\bibitem{stretch_light}
J.~D. Sandt, \emph{et~al.}, ``Stretchable optomechanical fiber sensors for
  pressure determination in compressive medical textiles,'' \emph{Advanced
  Healthcare Materials}, p. 1800293, Aug 2018.

\bibitem{Shan}
S.~Luo, \emph{et~al.}, ``{ViTac: Feature} sharing between vision and tactile
  sensing for cloth texture recognition,'' \emph{2018 IEEE International
  Conference on Robotics and Automation (ICRA)}, vol. abs/1802.07490, pp.
  2722--2727, 2018.

\bibitem{Lepora}
N.~Pestell, \emph{et~al.}, ``Dual-modal tactile perception and exploration,''
  \emph{IEEE Robotics and Automation Letters}, vol.~3, pp. 1033--1040, 2018.

\bibitem{Lepora_slip}
J.~W. James, N.~Pestell, and N.~F. Lepora, ``Slip detection with a biomimetic
  tactile sensor,'' \emph{IEEE Robotics and Automation Letters}, vol.~3, no.~4,
  pp. 3340--3346, Oct 2018.

\bibitem{ohka_takata_kobayashi_suzuki_morisawa_yussof_2009}
M.~Ohka, \emph{et~al.}, ``Object exploration and manipulation using a robotic
  finger equipped with an optical three-axis tactile sensor,'' \emph{Robotica},
  vol.~27, no.~5, pp. 763--770, 2009.

\bibitem{5306070}
K.~Sato, \emph{et~al.}, ``Finger-shaped gelforce: Sensor for measuring surface
  traction fields for robotic hand,'' \emph{IEEE Transactions on Haptics},
  vol.~3, no.~1, pp. 37--47, Jan 2010.

\bibitem{Nagano:78}
K.~Nagano, S.~Kawakami, and S.~Nishida, ``Change of the refractive index in an
  optical fiber due to externalforces,'' \emph{Applied Optics}, vol.~17,
  no.~13, pp. 2080--2085, Jul 1978.

\bibitem{Saga_reflection}
S.~Saga, R.~Taira, and K.~Deguchi, ``Precise shape reconstruction by active
  pattern in total-internal-reflection-based tactile sensor,'' \emph{IEEE
  Transactions on Haptics}, vol.~7, no.~1, pp. 67--77, Jan 2014.

\bibitem{Correll_boulder_silicone_force}
R.~Patel, R.~Cox, and N.~Correll, ``Integrated proximity, contact and force
  sensing using elastomer-embedded commodity proximity sensors,''
  \emph{Autonomous Robots}, Apr 2018.

\bibitem{Silicone_mechanical_property_Indentor}
H.~J. Qi, K.~Joyce, and M.~C. Boyce, ``Durometer hardness and the stress-strain
  behavior of elastomeric materials,'' \emph{Rubber Chemistry and Technology},
  vol.~76, no.~2, pp. 419--435, 2003.

\bibitem{balloon}
C.~H. King, \emph{et~al.}, ``Optimization of a pneumatic balloon tactile
  display for robot-assisted surgery based on human perception,'' \emph{IEEE
  Transactions on Biomedical Engineering}, vol.~55, no.~11, pp. 2593--2600, Nov
  2008.

\bibitem{pdms_materials}
Z.~Cai, \emph{et~al.}, ``A new fabrication method for all-{PDMS} waveguides,''
  \emph{Sensors and Actuators A: Physical}, vol. 204, pp. 44 -- 47, 2013.

\bibitem{kappassov_magnetic}
Z.~Kappassov, \emph{et~al.}, ``A series elastic tactile sensing array for
  tactile exploration of deformable and rigid objects,'' in \emph{2018 IEEE/RSJ
  International Conference on Intelligent Robots and Systems (IROS)}, Oct 2018,
  pp. 520--525.

\end{thebibliography}

\end{document}